\documentclass{article}

\usepackage{PRIMEarxiv}

\usepackage[utf8]{inputenc} 
\usepackage[T1]{fontenc}    
\usepackage{hyperref}       
\usepackage{url}            
\usepackage{booktabs}       
\usepackage{amsfonts}       
\usepackage{nicefrac}       
\usepackage{microtype}      
\usepackage{lipsum}
\usepackage{fancyhdr}       
\usepackage{graphicx}       
\graphicspath{{media/}}     
\usepackage{adjustbox}
\usepackage{amsmath}
\usepackage{float}

\title{RAGAT-Mind: A Multi-Granular Modeling Approach for Rumor Detection Based on MindSpore}

\author{
\begin{minipage}[t]{0.32\textwidth}
\centering
Zhenkai Qin$^{1,2,3}$\\
{\small $^1$School of Information Technology} \\
{\small $^2$Network Security Research Center} \\
{\small $^3$Big Data and Policing Technology Laboratory} \\
{\small Guangxi Police College, Nanning, China} \\
{\small \texttt{qinzhenkai@gxjcxy.edu.cn}}
\end{minipage}
\hfill
\begin{minipage}[t]{0.32\textwidth}
\centering
Guifang Yang \\
{\small School of Information Technology} \\
{\small Guangxi Police College} \\
{\small 530028, China} \\
{\small \texttt{yangguifang@gxjcxy.edu.cn}}
\end{minipage}
\hfill
\begin{minipage}[t]{0.32\textwidth}
\centering
Dongze Wu\\
{\small School of Information Technology} \\
{\small Guangxi Police College} \\
{\small 530028, China} \\
{\small \texttt{wudongze@gxjcxy.edu.cn}}
\end{minipage}
}

\begin{document}
\maketitle

\begin{abstract}

As false information continues to proliferate across social media platforms, effective rumor detection has emerged as a pressing challenge in natural language processing. This paper proposes \textbf{RAGAT-Mind}, a multi-granular modeling approach for Chinese rumor detection, built upon the MindSpore deep learning framework. The model integrates TextCNN for local semantic extraction, bidirectional GRU for sequential context learning, Multi-Head Self-Attention for global dependency focusing, and Bidirectional Graph Convolutional Networks (BiGCN) for structural representation of word co-occurrence graphs. Experiments on the Weibo1-Rumor dataset demonstrate that RAGAT-Mind achieves superior classification performance, attaining 99.2\% accuracy and a macro-F1 score of 0.9919. The results validate the effectiveness of combining hierarchical linguistic features with graph-based semantic structures. Furthermore, the model exhibits strong generalization and interpretability, highlighting its practical value for real-world rumor detection applications. 
\end{abstract}


\section{Introduction}

With the rapid development of social media, platforms such as Weibo, Twitter, and Facebook have become vital channels for information dissemination. These platforms offer users convenient content publishing tools and powerful diffusion mechanisms, enabling information to spread rapidly within a short period of time \cite{shu2017fake}. However, this highly efficient dissemination also provides fertile ground for the spread of misinformation, rumors, and erroneous public opinion. In scenarios such as emergencies, social hotspots, and public crises, the wide circulation of false content can easily mislead public perception and may even pose serious threats to social stability, policy implementation, and national security \cite{wu2015false,zhang2015character} .

In response to this issue, rumor detection has garnered increasing attention from both academia and industry. Traditional rumor detection methods are primarily based on handcrafted feature engineering, which involves extracting textual content, user behavior, and propagation graph structure, and then applying classifiers such as Support Vector Machines (SVM) and Random Forest to identify rumors\cite{cho2014learning}. Although these methods achieved promising results in the early stages, they heavily rely on domain expertise, exhibit poor generalizability, and struggle with the informal and diverse linguistic expressions prevalent in social media. Consequently, their robustness and adaptability are limited \cite{mu2023examining}.

With the widespread adoption of deep learning in natural language processing, an increasing number of studies have explored end-to-end rumor detection using neural networks. Convolutional Neural Networks (CNNs) excel at capturing local features and are well-suited for modeling n-gram patterns in text; Recurrent Neural Networks (RNNs), along with their variants such as Gated Recurrent Units (GRUs), are capable of capturing contextual dependencies, thereby enhancing the model’s understanding of textual sequences \cite{yang2019xlnet,yao2019graph} . In recent years, the introduction of the Transformer architecture and attention mechanisms—especially Multi-Head Self-Attention—has significantly improved performance in long-text and complex semantic modeling tasks, and has been widely applied in sentiment analysis, event detection, and related areas \cite{kipf2016semi} .

However, social media texts exhibit not only sequential characteristics but also explicit structural relationships, such as word co-occurrence and semantic association. Solely relying on sequence modeling often fails to capture the nonlinear structural dependencies between words. To address this limitation, Graph Neural Networks (GNNs) have been introduced into textual modeling tasks in recent years, enabling structure-aware representation learning. In particular, Bidirectional Graph Convolutional Networks (BiGCNs) leverage both forward and backward propagation to extract deep structural relationships from bidirectional semantic graphs. They have shown great potential in tasks such as text classification, sentiment analysis, and information extraction \cite{bian2020rumor} .

MindSpore is an all-scenario AI framework independently developed by Huawei. It supports hybrid execution of dynamic and static computational graphs, operator fusion optimization, and automatic parallelism, thereby enabling efficient training and deployment of deep neural networks. In Chinese natural language processing tasks, MindSpore provides flexible graph computation capabilities, operator support for heterogeneous structural modules, and efficient multi-module scheduling mechanisms. These features make it particularly suitable for complex modeling tasks that integrate convolutional, recurrent, attention-based, and graph-based neural architectures.

Building upon these foundations, this paper proposes a multi-granular modeling framework for Chinese rumor detection, named \textbf{RAGAT-Mind} (Rumor-aware Graph Attention and TextCNN model based on MindSpore), which achieves deep fusion between semantic and structural representation learning. The model integrates four key architectural modules: TextCNN for local semantic feature extraction, GRU for modeling temporal dependencies, Multi-Head Attention (MHA) for focusing on salient information, and a Bidirectional Graph Convolutional Network (BiGCN) for structural representation via word co-occurrence graphs. Through a parallel dual-path design, the model demonstrates strong expressiveness and robust discriminative capacity in handling Chinese social media rumors characterized by high linguistic ambiguity, fragmented structure, and frequent semantic discontinuities.

\section{Related Work}

With the rapid proliferation of social media platforms, the dissemination of information has become more efficient, but the spread of misinformation and online rumors has also intensified. As a result, automated rumor detection has emerged as a critical task in the field of natural language processing. Existing studies in this area primarily fall into three methodological categories: traditional machine learning approaches, deep semantic modeling methods, and graph-based neural modeling techniques.

Early research focused on handcrafted feature extraction combined with conventional classifiers. For instance, Rizzo et al.~\cite{rizzo2020adversarial} extracted statistical features from Weibo posts and integrated temporal propagation cues for early rumor detection. Shi et al.~\cite{shi2020artificial} applied transfer learning techniques to improve cross-domain adaptability among various Chinese datasets. Zhang et al.~\cite{zhang2021prioritizing} combined user behavioral graphs with textual features to estimate the credibility of information sources. Although these approaches provided valuable insights, they exhibited limited semantic expressiveness, inadequate structural modeling capabilities, and poor robustness to informal and ambiguous linguistic expressions prevalent in social media contexts.

With the advancement of deep learning, the field has shifted toward end-to-end representation learning. Convolutional Neural Networks (CNNs) have been widely adopted for capturing local n-gram semantics~\cite{hansen2019neural}, and Gated Recurrent Units (GRUs) have shown effectiveness in modeling temporal dependencies~\cite{krishnan2018adversarial}. Multi-Head Self-Attention (MHA), in particular, has improved the ability to focus on semantically salient regions~\cite{piciucco2021biometric}. Gao et al.~\cite{gao2022graph} introduced a sentiment-aware BERT framework to enhance the classification of fake news. However, most of these methods ignore the structural information embedded in text, which is essential for understanding the underlying topology of word associations.

To compensate for this deficiency, Graph Neural Networks (GNNs) have been introduced into rumor detection tasks to capture word co-occurrence and topological semantics. Zichao et al.~\cite{zhang2020every} proposed TextGCN to learn structure-aware representations by constructing heterogeneous word-document graphs. Yao et al.~\cite{yao2019graph} improved this framework by incorporating edge weighting and position encoding. Bian et al.~\cite{bian2020rumor} developed a Bidirectional Graph Convolutional Network (BiGCN) to model bidirectional semantic propagation. In the context of Chinese rumor detection, Vu et al.~\cite{vu2021rumor} proposed a multimodal fusion model combining textual features with user propagation paths via GCNs. Despite their effectiveness in capturing global structure, most GNN-based models rely on static topologies, making them insufficient for modeling dynamic semantic shifts and contextual variability in real-world discourse.

To bridge the gap between semantic and structural modeling, several fusion-based architectures have been proposed. Azri et al.~\cite{azri2021calling} presented a parallel CNN-GRU framework to jointly model local semantics and sequential dependencies. Wei et al.~\cite{wei2023dgtr} designed a dual-path model that integrates Transformer and GCN to capture both global semantics and topological structures. Mehta et al.~\cite{mehta2021transformer} incorporated Graph Attention Networks (GATs) into multi-head attention modules to perform semantic-aware node-level aggregation. Although these methods demonstrate promising results, they still face the following limitations:

\begin{itemize}
    \item \textbf{Static feature fusion:} Most fusion architectures rely on simple concatenation of outputs from different modules, lacking dynamic interactions or feedback mechanisms between semantic and structural branches.
    \item \textbf{Insufficient integration depth:} The coordination between convolutional, sequential, and graph components is often shallow, which limits the expressiveness of the joint representation.
    \item \textbf{Lack of adaptability to linguistic irregularities:} These methods struggle with semantic jumps, fragmented syntax, and informal language, which are prevalent in Chinese social media texts.
\end{itemize}

To address these challenges, we propose a unified multi-granular architecture---TextCNN-GRU-MHA-BiGCN---that seamlessly integrates convolutional, recurrent, attention-based, and graph-based modules. This hybrid design enhances both semantic expressiveness and structural awareness, offering a robust solution to the unique challenges of rumor detection in complex, dynamic, and linguistically diverse Chinese social media environments.

\section{ Model Description}
\subsection{Overall Architecture}

\begin{figure}[H]
\vspace{-0.5em} 
    \centering
    \includegraphics[width=0.95\textwidth]{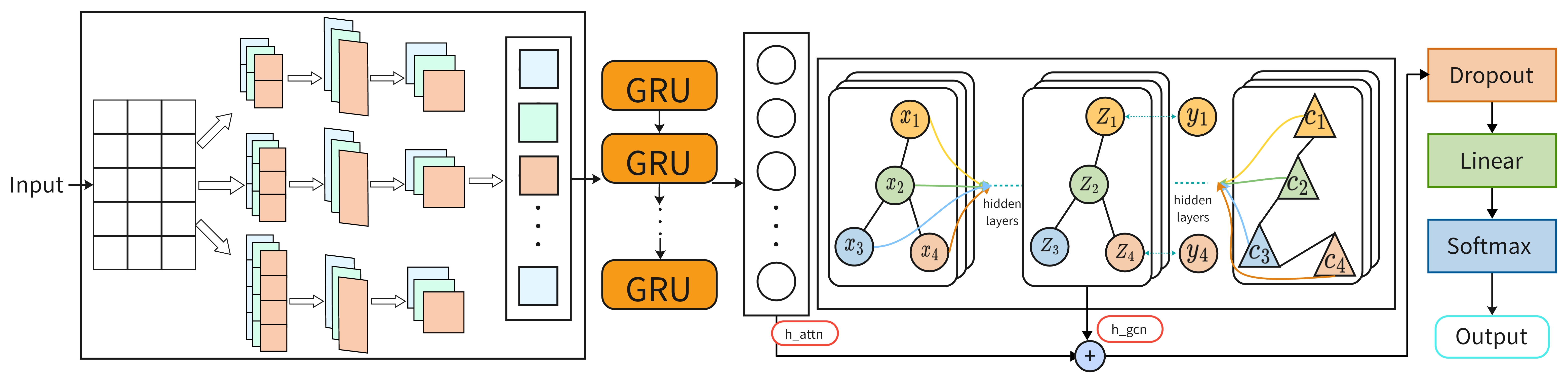}
    \caption{The overall architecture of the proposed TextCNN-GRU-MHA + BiGCN model.}
    \label{fig:architecture}
\vspace{-0.5em} 
\end{figure}

To effectively model the complex semantics and structural information present in Chinese social media rumors, we propose a multi-granular feature modeling framework named \textbf{RAGAT-Mind}, which integrates convolutional, recurrent, attention-based, and graph-based modules. As illustrated in Figure~\ref{fig:architecture}, the overall model consists of two parallel paths: a semantic modeling path and a structural modeling path. These two paths are designed to capture different levels of linguistic features, including local semantics, sequential dependencies, global attention, and word co-occurrence structures.

Each module in the architecture plays a specific role in the multi-granular representation learning process:
\begin{itemize}
  \item \textbf{TextCNN:} Extracts local n-gram semantic patterns and constructs initial phrase-level representations, which are particularly effective for capturing key expressions in short texts.
  \item \textbf{GRU:} Models temporal dependencies between words based on the convolutional features, enhancing sensitivity to contextual evolution.
  \item \textbf{MHA:} Employs a multi-head attention mechanism to assign higher weights to semantically important regions, improving global semantic modeling.
  \item \textbf{BiGCN:} Constructs a word co-occurrence graph and applies bidirectional graph convolution to capture structural dependencies beyond linear order.
  \item \textbf{Fusion Layer:} Concatenates the outputs from both semantic and structural paths, enabling unified modeling of linguistic content and latent structure.
\end{itemize}

In the semantic modeling path, the input text is first processed by the TextCNN module to extract multi-scale n-gram features, thereby constructing preliminary phrase-level representations. The convolutional filters slide over the word embedding matrix with various kernel sizes, capturing key phrase patterns within the sentence. These features are then fed into a Gated Recurrent Unit (GRU), which recursively encodes contextual relationships across time steps, effectively modeling long-range dependencies and semantic evolution. Given that GRU alone may not fully capture the varying importance of different semantic fragments, a Multi-Head Attention (MHA) mechanism is subsequently applied to aggregate token-level features with adaptive weighting. This enables the model to attend to task-relevant semantic regions and produce a context-aware global representation.

In parallel, the structural modeling path aims to capture non-linear dependencies between words. A word co-occurrence graph is constructed using a sliding window strategy, where each word is treated as a graph node, and edges are formed between words co-occurring within a local context window. This process yields a sentence-level adjacency matrix representing local syntactic or semantic structures. A Bidirectional Graph Convolutional Network (BiGCN) is then employed to perform graph convolutions on both the original and transposed adjacency matrices, thereby modeling bidirectional structural propagation and capturing both local and long-range co-occurrence patterns. This enhances the model’s ability to perceive non-sequential relationships such as word jumps, implicit pairings, and structural irregularities—common in Chinese social media rumor texts.

Finally, the outputs from the semantic and structural paths, namely the global semantic representation $\mathbf{h}{\text{attn}}$ and the structural embedding $\mathbf{h}{\text{gcn}}$, are concatenated to form a unified representation $\mathbf{h}_{\text{final}}$. This vector is passed through a dropout layer for regularization and then fed into a fully connected layer followed by a Softmax classifier for final binary prediction. Through the joint design of semantic and structural modules, RAGAT-Mind integrates local context modeling, long-range dependency capture, salient feature aggregation, and structural pattern extraction into a cohesive, interpretable, and robust rumor detection framework.

\subsection{TextCNN}
Chinese rumor texts on social media platforms are typically characterized by short sentence structures, high semantic density, and core information often conveyed through local phrases. These characteristics make convolutional structures particularly suitable for local semantic modeling. Therefore, this study adopts TextCNN as the first sub-module of the semantic modeling path. By applying one-dimensional convolutions with multiple receptive fields to the word embedding matrix, the model is able to extract n-gram semantic patterns at various granularities.

\begin{figure}[H]
    \centering
    \vspace{-0.5em} 
    \includegraphics[width=0.45\textwidth]{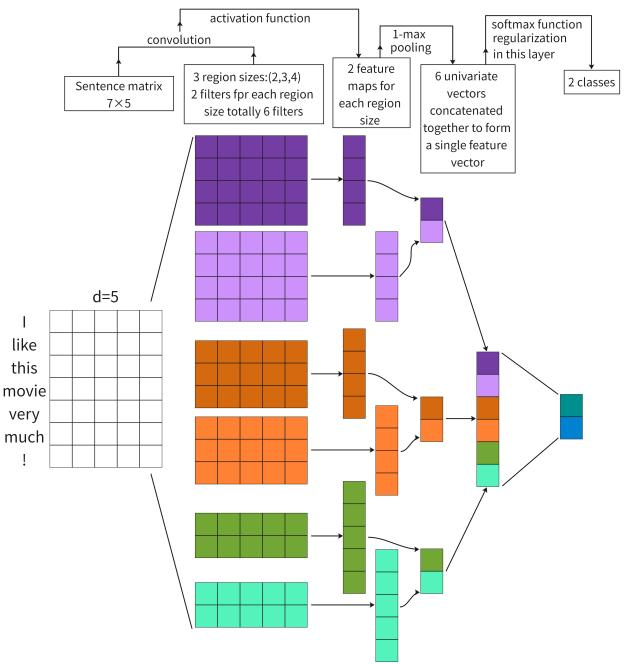}
    \caption{Structure of the TextCNN module}
    \label{fig:textcnn}

\end{figure}

As shown in Figure~\ref{fig:textcnn}, the TextCNN module consists of an embedding layer, multiple parallel one-dimensional convolutional filters, ReLU activation functions, max pooling layers, and a final feature concatenation structure. This architecture enables the model to extract semantic fragments of different lengths in parallel, significantly improving its sensitivity to local dependencies and providing high-quality input representations for downstream sequential modeling modules such as GRU.

In our implementation, we use three sets of convolutional filters with kernel sizes of 3, 4, and 5, corresponding to tri-gram to five-gram level semantic features. Each filter captures contextual combinations over different spans, enhancing the model’s ability to adapt to various granular phrase patterns. Considering that Chinese rumors on social platforms often contain informal language, short sentence structures, and concentrated information, semantics are usually conveyed through fragments rather than complete sentences. Relying on a single filter size may result in missed key combinations, limiting the model’s semantic coverage in short texts. To address this, we apply multiple receptive fields in parallel to capture diverse fragment-level semantic patterns and enhance the model's adaptability to non-standard expressions.Based on this design, the input to the TextCNN module is first mapped into a word embedding matrix $\mathbf{E} \in \mathbb{R}^{L \times d}$, where $L$ is the maximum sentence length and $d$ is the embedding dimension. For a one-dimensional convolutional filter $\mathbf{w}_i \in \mathbb{R}^{k \times d}$ of size $k$, the feature map is computed as follows:

\begin{equation} \mathbf{f}_i = \text{ReLU}(\mathbf{w}i * \mathbf{E}{i:i+k-1} + b_i) \end{equation}

Each convolutional output is then passed through a max pooling operation to extract the most salient local response:

\begin{equation} \text{MaxPool}(\mathbf{f}_i) = \max(\mathbf{f}_i) \end{equation}

To integrate semantic features from different scales, the pooled outputs from all convolutional filters are concatenated into a unified semantic vector:

\begin{equation} \mathbf{h}_{\text{CNN}} = [\text{MaxPool}(\mathbf{f}_1); \text{MaxPool}(\mathbf{f}_2); \ldots; \text{MaxPool}(\mathbf{f}_K)] \end{equation}

The resulting vector $\mathbf{h}_{\text{CNN}}$ encapsulates local semantic signals extracted across multiple receptive fields and serves as the input for subsequent sequential modeling modules.

\subsection{GRU}
Although convolutional structures excel at capturing local n-gram semantic features, their fixed receptive fields limit the ability to model long-range dependencies between words. To address this limitation, we incorporate a Gated Recurrent Unit (GRU) after the TextCNN module to capture temporal dependencies within the sequence and enhance the dynamic modeling of contextual semantic evolution.

\begin{figure}[H]
\vspace{-0.3em}
    \centering
    \includegraphics[width=0.55\textwidth]{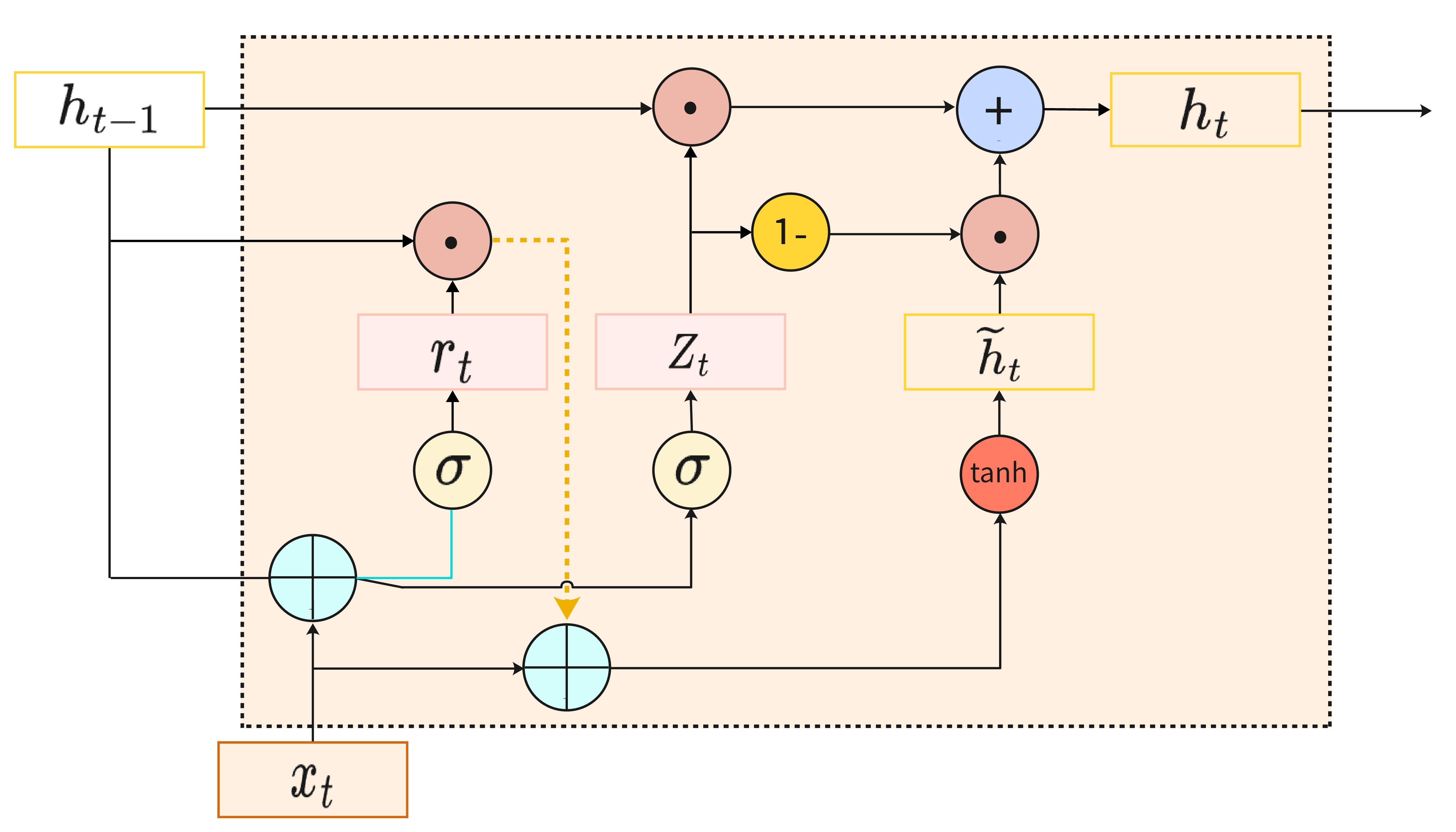}
    \caption{Structure of the GRU module}
    \label{fig:gru}
\end{figure}

In the proposed architecture, the GRU module serves as a core component of the semantic modeling path, as illustrated in Figure~\ref{fig:gru}. It receives multi-scale local features extracted by the TextCNN module and performs recursive modeling along the temporal dimension to capture dynamic dependencies among tokens. Building on the local representations provided by convolution, the GRU integrates historical and current information via gating mechanisms, transforming static features into dynamic semantic representations with contextual coherence. This process significantly enhances the model’s capacity to capture long-distance dependencies while suppressing redundant and noisy inputs, thereby improving the robustness and stability of semantic modeling.

Formally, let the output feature sequence of the TextCNN be denoted as $\mathbf{h}{\text{CNN}} \in \mathbb{R}^{L \times C}$, where $L$ is the sequence length and $C$ is the number of channels. At time step $t$, the hidden state of the GRU is updated based on the current input $\mathbf{h}{\text{CNN}, t}$ and the previous hidden state $\mathbf{h}_{t-1}$ as follows:

\begin{equation} \mathbf{h}t = \text{GRU}(\mathbf{h}{\text{CNN}, t}, \mathbf{h}_{t-1}) \end{equation}

Internally, the GRU dynamically regulates the flow of information using update and reset gates:

\begin{align} z_t &= \sigma(W_z \mathbf{x}t + U_z \mathbf{h}{t-1}) \ r_t &= \sigma(W_r \mathbf{x}t + U_r \mathbf{h}{t-1}) \end{align}

Guided by these gating mechanisms, the candidate hidden state is computed as:

\begin{equation} \tilde{\mathbf{h}}_t = \tanh(W_h \mathbf{x}t + U_h (r_t \odot \mathbf{h}{t-1})) \end{equation}

The final hidden state is then updated via a gated combination of the previous state and the candidate state:

\begin{equation} \mathbf{h}t = (1 - z_t) \odot \mathbf{h}{t-1} + z_t \odot \tilde{\mathbf{h}}_t \end{equation}

The sequence of hidden states output by the GRU, denoted as $\mathbf{H}_{\text{GRU}}$, serves as the input to the subsequent Multi-Head Attention (MHA) module. This output provides a semantically distinct and temporally coherent representation for modeling global dependencies. Compared to directly feeding convolutional features into the attention mechanism, the intermediate sequential modeling provided by the GRU significantly improves the precision of attention focus and the model’s overall awareness of contextual semantics.

\subsection{MHA}

Although the Gated Recurrent Unit (GRU) demonstrates considerable effectiveness in modeling temporal dependencies within word sequences, its reliance on sequential order inherently limits its capacity to capture long-range semantic interactions between non-adjacent tokens. This limitation becomes particularly salient in the context of Chinese social media rumors, where non-linear linguistic structures—such as incomplete sentence patterns, disordered syntax, and semantic discontinuities—are frequently observed. Sequential modeling mechanisms, such as GRU, often struggle to accommodate these irregularities, thereby impeding comprehensive semantic understanding and reducing classification precision. To address this issue, we introduce a Multi-Head Attention (MHA) mechanism following the GRU layer. This module enhances the model’s ability to capture global semantic dependencies and effectively encodes long-distance relationships, thereby mitigating the limitations of purely sequential models in processing non-standard textual structures.

In the proposed framework, the MHA module plays a critical role in the semantic modeling path by integrating global contextual information. On one hand, it inherits the contextual coherence constructed by the GRU; on the other hand, it simultaneously models non-local interactions across the sequence through multiple attention heads, enabling multi-perspective semantic fusion. The attention weight matrix explicitly reflects the relative importance of each word within the sentence, guiding the model to focus on the most informative regions for rumor detection while suppressing irrelevant or noisy signals. This mechanism ultimately contributes to improved classification accuracy and robustness.

Technically, the MHA module takes the GRU-generated hidden state sequence $\mathbf{H}_{\text{GRU}} \in \mathbb{R}^{L \times d}$ as input and applies three parallel linear transformations to compute the query ($\mathbf{Q}$), key ($\mathbf{K}$), and value ($\mathbf{V}$) matrices:

\begin{equation} \mathbf{Q} = \mathbf{H}{\text{GRU}} W^Q, \quad \mathbf{K} = \mathbf{H}{\text{GRU}} W^K, \quad \mathbf{V} = \mathbf{H}_{\text{GRU}} W^V \end{equation}

where $W^Q$, $W^K$, and $W^V \in \mathbb{R}^{d \times d_k}$ are trainable projection matrices, and $d_k$ denotes the dimensionality of each attention head. The attention distribution is then computed using the scaled dot-product mechanism:

\begin{equation} \text{Attention}(\mathbf{Q}, \mathbf{K}, \mathbf{V}) = \text{Softmax}\left(\frac{\mathbf{QK}^\top}{\sqrt{d_k}}\right) \mathbf{V} \end{equation}

This operation is conducted in parallel across $h$ attention heads to capture diverse semantic relationships in different subspaces. The outputs of all heads are concatenated and projected back to the original dimensionality via a final linear transformation:

\begin{equation} \text{MHA}(\mathbf{H}_{\text{GRU}}) = \text{Concat}(\text{head}_1, \ldots, \text{head}_h) W^O \end{equation}

where $W^O \in \mathbb{R}^{hd_k \times d}$ is the output projection matrix, and $\text{head}_i$ denotes the output from the $i$-th attention head.

\subsection{ BiGCN}

Although the aforementioned semantic modeling path (TextCNN--GRU--MHA) demonstrates promising capabilities in capturing local semantics, temporal evolution, and global dependencies within text, its architecture remains fundamentally grounded in a linear sequential structure, which limits its ability to explicitly represent latent dependencies between non-adjacent words. To address this limitation, we introduce a structural modeling branch based on a Bidirectional Graph Convolutional Network (BiGCN), which aims to model topological dependencies among words through the lens of graph neural networks. This module provides a complementary structural representation distinct from sequential modeling, enhancing the model's robustness in complex texts characterized by structural irregularities and semantic jumps.

Specifically, a word co-occurrence graph is constructed for each input sentence using a sliding window mechanism. Each word is treated as a graph node, and co-occurring word pairs within the window are connected by edges, yielding an adjacency matrix $\mathbf{A} \in \mathbb{R}^{L \times L}$, where $L$ denotes the sentence length. This structure encodes local word-level associations and serves as a structural prior for subsequent graph convolution operations.

Based on the constructed graph, BiGCN performs convolutions on both the original adjacency matrix $\mathbf{A}$ and its transpose $\mathbf{A}^\top$ to model bidirectional semantic propagation. Let $\mathbf{X} \in \mathbb{R}^{L \times d}$ denote the input word embeddings, the forward and backward propagation can be formulated as:

\begin{equation}
\mathbf{H}^{f} = \text{ReLU}(\mathbf{A} \mathbf{X} \mathbf{W}^{f}), \quad 
\mathbf{H}^{b} = \text{ReLU}(\mathbf{A}^\top \mathbf{X} \mathbf{W}^{b})
\end{equation}

where $\mathbf{W}^{f}$ and $\mathbf{W}^{b}$ are learnable weight matrices for forward and backward propagation, respectively, and $\mathbf{H}^{f}$ and $\mathbf{H}^{b}$ denote the resulting features. This bidirectional mechanism allows for symmetrical structural aggregation, mitigating the limitations of unidirectional graph propagation.

The forward and backward features are concatenated to form the comprehensive structural representation:

\begin{equation}
\mathbf{H}^{\text{BiGCN}} = [\mathbf{H}^{f}; \mathbf{H}^{b}]
\end{equation}

To derive a sentence-level representation, mean pooling is applied across all nodes:

\begin{equation}
\mathbf{h}_{\text{gcn}} = \text{MeanPool}(\mathbf{H}^{\text{BiGCN}})
\end{equation}

Finally, the output from the structural path $\mathbf{h}_{\text{gcn}}$ is fused with the semantic representation $\mathbf{h}_{\text{attn}}$ in the subsequent fusion module, providing multi-granular, structurally-aware features to support robust and accurate rumor classification.

\section{Experimental Setup and Results}

\subsection{Dataset Description}
We conduct experiments using the publicly available Chinese Weibo rumor detection dataset, Weibo1-Rumor, which consists of 3,387 real-world social media texts, including 1,538 rumor instances and 1,849 non-rumors. All texts are collected from actual online platforms, reflecting realistic linguistic characteristics. To ensure reproducibility and generalization, the dataset is split into training and testing sets at an 8:2 ratio. Preprocessing steps include word segmentation, token indexing, and adjacency matrix construction.

\subsection{Experimental Environment}

All experiments in this study were conducted on a Windows 10 operating system using the MindSpore deep learning framework developed by Huawei. The hardware environment was configured with an Intel\textsuperscript{\textregistered} Xeon\textsuperscript{\textregistered} CPU, 64 GB of RAM, and an NVIDIA GeForce RTX 3070 Laptop GPU, providing sufficient computational resources and high-throughput performance during model training. MindSpore supports both dynamic and static graph execution modes, and offers efficient operator fusion and heterogeneous computing capabilities, making it well-suited for large-scale text modeling tasks. The software environment was built upon Python 3.8 and integrated several essential libraries, including \texttt{pandas}, \texttt{numpy}, \texttt{jieba}, \texttt{scikit-learn}, and \texttt{tqdm}, which were employed for data preprocessing, Chinese word segmentation, evaluation metric computation, and training progress visualization, respectively.

During training, the Adam optimizer was utilized in conjunction with a dynamic learning rate decay strategy to accelerate convergence and improve model stability. Dropout regularization was introduced at various layers to mitigate overfitting. All experiments were conducted under a unified configuration with fixed random seed settings to ensure fairness and reproducibility. Moreover, the Early Stopping strategy was enabled to automatically terminate training if validation performance failed to improve for several consecutive epochs, thereby enhancing the model’s generalization capacity and optimizing resource utilization.

The key hyperparameter settings used in the experiments are summarized in Table~\ref{tab:hyperparams}.
\begin{table}[H]
\vspace{-0.8em}
\centering
\setlength{\abovecaptionskip}{2pt}
\setlength{\belowcaptionskip}{2pt}

\caption{Model Hyperparameter Settings}
\label{tab:hyperparams}

\renewcommand{\arraystretch}{1.2}
\setlength{\tabcolsep}{14pt}

\begin{tabular}{p{6cm}p{3cm}} 
\toprule
\textbf{Hyperparameter} & \textbf{Value} \\
\midrule
Max sequence length & 128 \\
Embedding dimension & 128 \\
Convolution kernel sizes & [3, 4, 5] \\
GRU hidden size & 128 \\
Number of attention heads & 4 \\
Dropout rate & 0.5 \\
Batch size & 32 \\
Epochs & 3 \\
Learning rate & 0.001 \\
Optimizer & Adam \\
\bottomrule
\end{tabular}
\end{table}


\subsection{Evaluation Metrics}
To comprehensively evaluate the classification performance of the proposed model on the Chinese rumor detection task, four commonly used metrics are employed: Accuracy, Precision, Recall, and F1-Score. Accuracy reflects the overall correctness of predictions; Precision measures the proportion of correctly predicted positive instances among all predicted positives; Recall indicates the proportion of actual positive instances that are correctly identified; and F1-Score serves as a harmonic mean of Precision and Recall, providing a balanced assessment of the model’s robustness and discriminative capability. To reduce the potential bias caused by class imbalance, we adopt a macro-averaging strategy, where Precision, Recall, and F1 are calculated independently for each class and then averaged to ensure that each class contributes equally to the final evaluation. All metrics are computed using the \texttt{classification\_report} function from the \texttt{scikit-learn} package, ensuring the accuracy and reproducibility of the evaluation process.
\begin{equation}
\text{Accuracy} = \frac{TP + TN}{TP + TN + FP + FN}
\end{equation}

\begin{equation}
\text{Precision} = \frac{TP}{TP + FP}
\end{equation}

\begin{equation}
\text{Recall} = \frac{TP}{TP + FN}
\end{equation}

\begin{equation}
\text{F1} = 2 \times \frac{\text{Precision} \times \text{Recall}}{\text{Precision} + \text{Recall}}
\end{equation}

Here, $TP$ denotes the number of true positives, $TN$ the number of true negatives, $FP$ the number of false positives, and $FN$ the number of false negatives.

\subsection{Experimental Results and Analysis}

To comprehensively evaluate the performance of the proposed RAGAT-Mind model in Chinese social media rumor detection, we conducted comparative experiments under a unified experimental environment and consistent dataset partitioning. Several representative baseline models were selected for comparison, including: TextCNN, which focuses on local semantic feature extraction; GRU-ATT, which integrates sequential modeling with attention mechanisms; TextGCN, which emphasizes structural representation based on word co-occurrence graphs; BERT-FT, a contextual model built on large-scale pretraining; and GAT, which introduces node-level attention within graph structures. All models were trained and evaluated on the Weibo1-Rumor dataset. Table~\ref{tab:comparison_extended} summarizes the overall performance of each model in terms of Accuracy, Precision, Recall, Macro-F1, and inference latency (i.e., time per sample).
\begin{table}[H]
\centering
\caption{Extended performance comparison on the Weibo1-Rumor test set}
\label{tab:comparison_extended}
\vspace{-0.5em}
\renewcommand{\arraystretch}{1.2}
\setlength{\tabcolsep}{10pt}
\begin{tabular}{lccccc}
\toprule
\textbf{Model} & \textbf{Accuracy (\%)} & \textbf{Precision} & \textbf{Recall} & \textbf{Macro-F1} & \textbf{Time (ms)} \\
\midrule
TextCNN        & 95.82   & 0.9471   & 0.9630   & 0.9546   & 3.12  \\
GRU-ATT        & 96.17   & 0.9558   & 0.9651   & 0.9603   & 4.89  \\
TextGCN        & 97.06   & 0.9630   & 0.9797   & 0.9712   & 6.71  \\
BERT-FT        & 98.12   & 0.9801   & 0.9765   & 0.9783   & 12.34 \\
GAT            & 97.54   & 0.9694   & 0.9832   & 0.9762   & 9.43  \\
\textbf{RAGAT-Mind (Ours)} & \textbf{99.20} & \textbf{0.9925} & \textbf{0.9914} & \textbf{0.9919} & \textbf{5.36} \\
\bottomrule
\end{tabular}
\vspace{-0.9em}
\end{table}
As shown in the table, RAGAT-Mind significantly outperforms all baseline models in terms of both Accuracy and Macro-F1, demonstrating consistent and robust performance. During training, the model's accuracy improved from 89.6\% to 99.2\%, while the Macro-F1 score increased from 0.8939 to 0.9919. Meanwhile, the training loss decreased from 0.5896 to 0.0447, indicating excellent convergence behavior and generalization capability. Compared with the base model TextCNN, RAGAT-Mind achieved improvements of approximately 3.4\% in accuracy and 3.7\% in Macro-F1. Even in comparison with structurally aware models such as TextGCN, RAGAT-Mind achieved a 2.14\% gain in accuracy, highlighting the effectiveness of its structural-semantic joint modeling strategy.

From the perspective of modeling mechanisms, TextCNN is effective at capturing local n-gram features but lacks the ability to model long-range dependencies. GRU-ATT introduces attention to enhance sequential context modeling, yet it is still limited in terms of structural awareness. TextGCN and GAT capture global co-occurrence relationships via graph structures, but they lack temporal modeling capabilities for semantic evolution. In contrast, RAGAT-Mind adopts a parallel dual-path design integrating TextCNN, GRU, Multi-Head Attention (MHA), and BiGCN modules, collectively supporting local semantics, sequential dependencies, salient region focusing, and global structural modeling. This architectural synergy greatly enhances the model's adaptability to complex semantic patterns in Chinese social media texts.

\begin{figure}[H]
\vspace{-1.0em}
    \centering
    \includegraphics[width=0.70\textwidth]{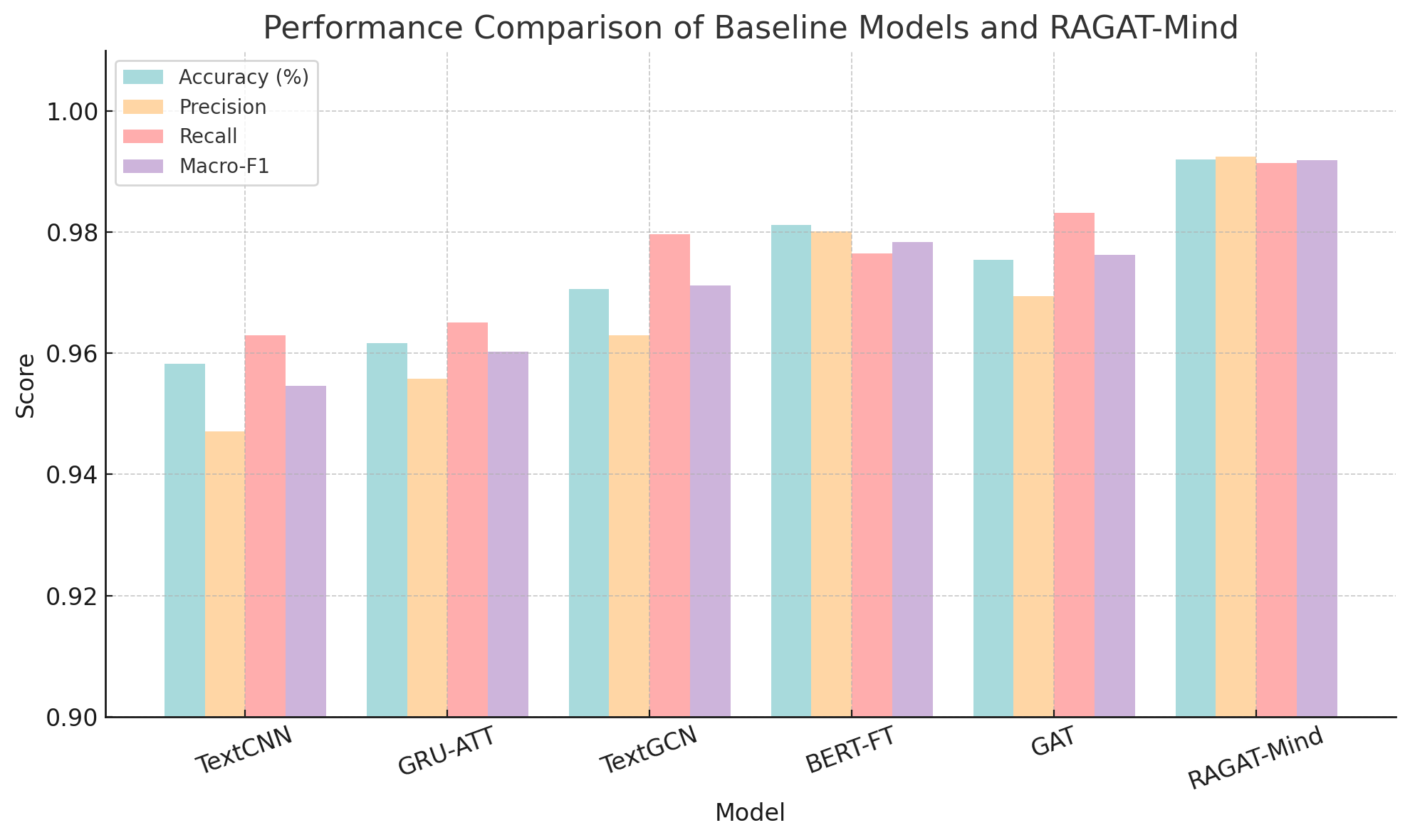}
\vspace{-0.8em}
    \caption{Performance Comparison Across All Models}
    \label{fig:bar_comparison_soft}
\vspace{-0.8em}
\end{figure}

To provide a clearer comparison of performance across all models, Figure~\ref{fig:bar_comparison_soft} presents a bar chart comparing the four key metrics: Accuracy, Precision, Recall, and Macro-F1. As illustrated, traditional sequence models such as TextCNN and GRU-ATT exhibit relatively strong performance in Precision and Recall, reflecting their effectiveness in capturing explicit semantic features. However, their Macro-F1 scores are notably lower in imbalanced or linguistically irregular cases, suggesting limited generalizability and blurry classification boundaries. Graph-based methods like TextGCN and GAT achieve notable gains in Recall, validating the utility of structural modeling for discovering latent semantic structures. Nevertheless, their lack of temporal modeling constrains their ability to capture evolving contextual semantics, which in turn affects their overall accuracy.

Further analysis reveals that BERT-FT performs well across all metrics, benefiting from deep contextual embeddings and prior knowledge from pretraining. However, its relatively high inference latency (12.34 ms) imposes limitations on deployment in latency-sensitive applications. In contrast, RAGAT-Mind achieves the best performance across all metrics—Accuracy (99.20\%), Precision (0.9925), Recall (0.9914), and Macro-F1 (0.9919)—while maintaining low inference latency (5.36 ms). The uniform bar heights in Figure~\ref{fig:bar_comparison_soft} reflect the model’s consistent robustness and discriminative capability when handling noisy, fragmented, and semantically diverse rumor texts. These results further validate the effectiveness of the model’s hybrid semantic-structural design.

In summary, RAGAT-Mind demonstrates high stability, strong discriminative accuracy, and structural sensitivity in Chinese social media rumor detection tasks. The experimental findings fully support the model’s multi-granular representation learning framework and provide a robust technical foundation for future research in complex and adversarial linguistic environments.

\section{Discussion}

To further explore the applicability and potential of the proposed model for Chinese rumor detection, this section provides an in-depth discussion of the experimental outcomes. The analysis is conducted from three perspectives: classification performance and model limitations, interpretability and structural understanding, and practical adaptability and future research directions.

\subsection{Classification Effectiveness and Model Constraints}

Experimental results show that the proposed TextCNN-GRU-MHA-BiGCN model achieves strong classification performance, attaining an accuracy of 99.2\% and a macro-average F1 score of 0.9919, with a low loss value of 0.0447. These results demonstrate the model’s ability to fit and generalize effectively across rumor classification tasks. The success lies in the integration of four complementary mechanisms: local semantic feature extraction (TextCNN), sequential dependency modeling (GRU), attention-based key region focusing (MHA), and structural representation learning (BiGCN). Together, these components provide a comprehensive representation of short-text semantics, enhancing the model’s robustness against expression diversity.

However, several limitations remain. First, model performance is highly sensitive to the quality of training data. Semantic ambiguity and noisy labels in social media datasets may interfere with the learning process and degrade generalizability. Second, the current graph structure is built on a sliding window co-occurrence mechanism, which fails to incorporate higher-level linguistic features such as syntactic dependencies, semantic roles, or external knowledge. This restricts the model’s capacity to capture deeper language structures. Lastly, the model has not been validated across cross-platform or cross-domain scenarios. Its ability to adapt to heterogeneous data from different social platforms remains untested and requires further investigation. While the model performs well on a single-source dataset, future deployment in real-world applications demands enhancements in domain adaptability and resilience to data diversity.

\subsection{Interpretability and Structural Awareness}

In addition to achieving strong classification results, the proposed model emphasizes architectural interpretability and structural transparency. The multi-head self-attention mechanism enables dynamic weighting of token-level semantic contributions by calculating attention scores across word pairs. These scores can be visualized to identify key semantic regions that influence classification decisions, offering insight into the model's decision-making process. Furthermore, the BiGCN module captures structural dependencies in both forward and backward directions within the word graph. This dual-propagation mechanism enhances the model’s sensitivity to hierarchical linguistic structures and latent relationships between words.

The final feature fusion process concatenates semantic and structural representations, preserving their individual contributions and enabling interpretability at both the content and structure levels. This design facilitates post-hoc model diagnosis and interpretability analysis, making the framework more transparent and trustworthy in real-world applications. Overall, the architecture provides not only accurate predictions but also a foundation for interpretable decision-making, which is increasingly important for high-stakes applications such as misinformation detection.

\subsection{Practical Adaptability and Future Directions}

Although the model demonstrates strong performance on the Weibo1-Rumor dataset, broader practical deployment requires further generalization and adaptability improvements. First, the current experimental setup focuses exclusively on a single platform (Weibo), without incorporating datasets from other popular Chinese platforms such as Zhihu or WeChat public accounts. Future work should evaluate the model’s robustness across platforms by employing domain adaptation and transfer learning strategies to address distribution shifts.

Second, the current graph construction relies solely on window-based co-occurrence, which may be insufficient for capturing complex syntactic or knowledge-based relations. Integrating external linguistic resources—such as dependency parsing, knowledge graphs, or named entity relations—could significantly enhance the expressiveness and contextual grounding of the constructed graphs.

Finally, in real-world scenarios, rumor propagation often involves multimodal elements, including text-image combinations and behavioral propagation paths. Future extensions of this work could involve designing multimodal architectures that incorporate visual, behavioral, or temporal signals into the framework, further enhancing detection robustness and real-world usability.

\section{Conclusion}

This study presents a multi-module integrated model for Chinese rumor detection that combines convolutional, recurrent, attention-based, and graph-based neural architectures. Specifically, the model incorporates a Text Convolutional Neural Network (TextCNN) for local n-gram pattern extraction, a Gated Recurrent Unit (GRU) for sequential dependency modeling, Multi-Head Self-Attention (MHA) for enhancing focus on semantically important regions, and a Bidirectional Graph Convolutional Network (BiGCN) for capturing structural dependencies via word co-occurrence graphs. The integration of these complementary components enables the model to jointly model local semantics, global context, and graph-based structure within a unified framework.

Comprehensive experiments conducted on the Weibo1-Rumor dataset demonstrate the model’s effectiveness and stability, achieving a test accuracy of 99.2\% and a macro-average F1 score of 0.9919. These results confirm the superiority of the proposed hybrid architecture in short-text classification tasks and validate the value of multi-perspective representation learning in complex social media environments. Furthermore, the model’s design preserves structural interpretability, providing both performance and explainability, which are critical for misinformation detection in real-world applications.

Despite its strong performance, the model also presents opportunities for further improvement. Future work will focus on enhancing the semantic depth of graph construction by incorporating syntactic dependencies or external knowledge structures such as entity relations or knowledge graphs. In addition, transfer learning and domain adaptation techniques will be explored to improve generalizability across diverse social media platforms. Finally, extending the model to multimodal rumor detection by incorporating visual and behavioral cues is expected to further boost its robustness and application potential.

Beyond methodological contributions, RAGAT-Mind offers valuable potential for real-world deployment. The model can be integrated into content monitoring systems for governmental platforms, social media portals, or public safety dashboards to enable real-time rumor screening and content credibility assessment. Its unified, interpretable architecture ensures both accuracy and transparency, making it suitable for applications where both reliability and explainability are critical. Overall, the proposed model contributes a scalable, interpretable, and high-performance framework for advancing the state of the art in rumor detection research and practical deployment.

\section*{Acknowledgments}
Thanks for the support provided by the MindSpore Community.
\bibliographystyle{unsrt}  
\bibliography{references}

\end{document}